\documentclass[letterpaper]{article} 

\usepackage{aaai25}  
\usepackage{times}  
\usepackage{helvet}  
\usepackage{courier}  
\usepackage[hyphens]{url}  
\usepackage{graphicx} 
\usepackage{natbib}  
\usepackage{caption} 
\usepackage{algorithm}
\usepackage{algorithmic}
\usepackage{subcaption}
\usepackage{listings}
\usepackage{xcolor}

\usepackage{newfloat}
\usepackage{listings}

\DeclareCaptionStyle{ruled}{labelfont=normalfont,labelsep=colon,strut=off} 
\floatstyle{ruled}
\newfloat{listing}{tb}{lst}{}
\floatname{listing}{Listing}
\title{Atari-GPT: Benchmarking Multimodal Large Language Models as Low-Level Policies in Atari Games}
\author{
    Nicholas R. Waytowich\textsuperscript{\rm 1},
    Devin White\textsuperscript{\rm 2},
    MD Sunbeam\textsuperscript{\rm 2},
    Vinicius G. Goecks\textsuperscript{\rm 1}
}
\affiliations{
    \textsuperscript{\rm 1}DEVCOM Army Research Laboratory\\
    \textsuperscript{\rm 2} Army Educational Outreach Program\\
    nicholas.r.waytowich.civ@army.mil
}

\begin{document}

\maketitle

\begin{abstract}
Recent advancements in large language models (LLMs) have expanded their capabilities beyond traditional text-based tasks to multimodal domains, integrating visual, auditory, and textual data. While multimodal LLMs have been extensively explored for high-level planning in domains like robotics and games, their potential as low-level controllers remains largely untapped. In this paper, we introduce a novel benchmark aimed at testing the emergent capabilities of multimodal LLMs as low-level policies in Atari games. Unlike traditional reinforcement learning (RL) methods that require training for each new environment and reward function specification, these LLMs utilize pre-existing multimodal knowledge to directly engage with game environments. Our study assesses the performances of multiple multimodal LLMs against traditional RL agents, human players, and random agents, focusing on their ability to understand and interact with complex visual scenes and formulate strategic responses. Our results show that these multimodal LLMs are not yet capable of being zero-shot low-level policies. Furthermore, we see that this is, in part, due to their visual and spatial reasoning. Additional results and videos are available on our project webpage: \url{https://dev1nw.github.io/atari-gpt/}.
\end{abstract}

\section{Introduction}
\label{sec:intro}  

\begin{figure*}[ht]
\begin{center}
\centerline{\includegraphics[width=0.85\textwidth]{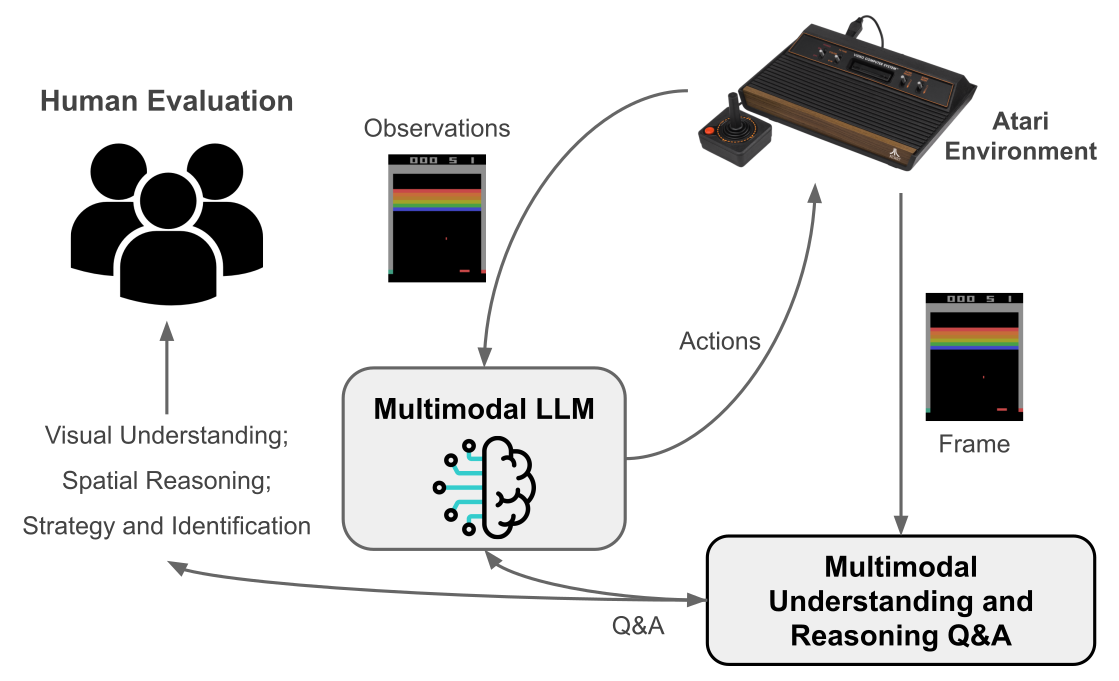}}
\caption{\textbf{Atari-GPT: System diagram:} illustrates the integration of a multimodal large language model (LLM) as a low-level agent within the Atari gaming environment. It highlights the flow of inputs from the game to the LLM and back, demonstrating how the model processes game observations and generates corresponding actions. Additionally, the diagram includes the framework for human evaluation, which assesses the LLM's capabilities in visual understanding, spatial reasoning, strategic intuition, and environment recognition through a structured Q\&A process.}
\label{fig:Atari_diagram}
\end{center}
\end{figure*}

Advancements in natural language processing, dataset scaling, and model scaling have led to large language models, specifically ChatGPT (GPT-3.5) \cite{ChatGPT}, which revolutionized text-to-text models. Evolving from these models are more advanced multimodal models with the ability to take multiple types of input like text, images, and even audio, like GPT-4o and Gemini \cite{openai2024gpt4,reid2024gemini,gpt4o}. In addition, with each new iteration of these large multimodal models, we see vast improvements in efficiency. For example, the development of GPT-4 Turbo to GPT-4o to GPT-4o mini highlights the case where sacrificing slight general capabilities improves the inference cost and speed \cite{gpt4omini}.

With each development of these multimodal models, they show potential beyond their traditional conversational task. Researchers have investigated their capabilities in areas like robotics and high-level planning in automated systems~\cite{li2023interactive, rana2023sayplan}. However, much of the current literature focuses on utilizing multimodal models for high-level planning \cite{xu2024surveyroboticsfoundationmodels}, leaving their use as low-level controllers unexplored, akin to what is typically learned by reinforcement learning agents in complex environments like video games.

To investigate whether multimodal LLMs can function effectively as low-level controllers, we perform initial tests on GPT-4V~\cite{openai2024gpt4}, GPT-4o~\cite{gpt4o}, Gemini Flash~\cite{geminiFlash}, and Claude 3 Haiku~\cite{Claude3} in Atari. Along with the raw performance of each of these models, we investigate their visual understanding, spatial reasoning, and strategy formulation across multiple environments.


In this paper, we show that these multimodal models are not yet capable of zero-shot game-play in Atari. We found that this is, in part, due to their inability to understand the visual and spatial components of a given game-play image. We do this by introducing a novel benchmark for multimodal LLMs to explore their emergent capabilities as low-level policies in Atari games as outlined in Figure \ref{fig:Atari_diagram}.

\section{Atari-GPT}
\label{sec:atarigpt}

We present a set of experiments designed to benchmark the effectiveness of multimodal LLMs as low-level decision-making agents in the domain of Atari video games, which we refer to as \textbf{``Atari-GPT"}. Our primary focus is assessing the models' game-playing capabilities and performance measured by several factors: the game score, visual understanding, spatial reasoning, and proficiency in devising efficient game strategies. 

First, we evaluate the multimodal LLMs' performance in playing Atari as a low-level policy, judged by each game's score. This assessment measures the models' success by comparing their performance to standard reinforcement learning algorithms, random agents, and human players, analyzing how well the models can act as low-level policies by making decisions based on the current game state.

Second, we examine the multimodal LLMs' visual understanding and spatial reasoning capabilities. We do this by testing how well the models properly identify different key visual elements within a given frame, understand how these elements are related to one another spatially, and the ability of the models to create a meaningful strategy based on their scene understanding. Additionally, we test if the models are able to properly identify the game environment when given no context other than the image.
For testing visual understanding and spatial reasoning, we use the same set of Atari environments used to evaluate game-play performance with the addition of another environment, Basic Math.

This experimental structure provides a more comprehensive analysis of the decision-making processes of LLMs by assessing their overall understanding of the game environment within Atari video games, and evaluating their performance as low-level policies. Through this methodology, we aim to establish a new benchmark for evaluating LLMs in low-level control tasks, exploring how these language models compare to humans and learning algorithms.

\section{Experimental Setup}



\begin{figure*}[t!]
     \centering
     \begin{subfigure}[b]{0.5\columnwidth}
         \centering
         \includegraphics[width=\columnwidth]{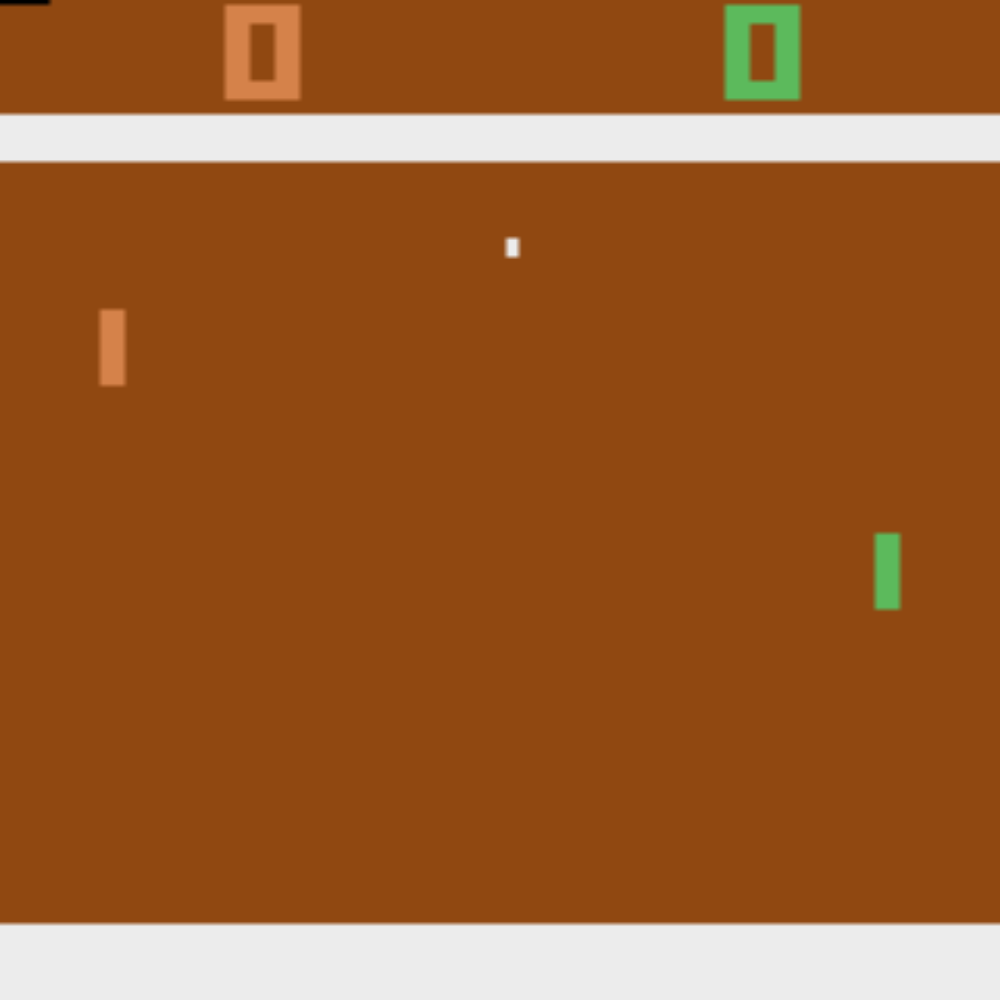}
         \caption{Pong}
         \label{fig:pong}
     \end{subfigure}
     \hfill
     \begin{subfigure}[b]{0.5\columnwidth}
         \centering
         \includegraphics[width=\columnwidth]{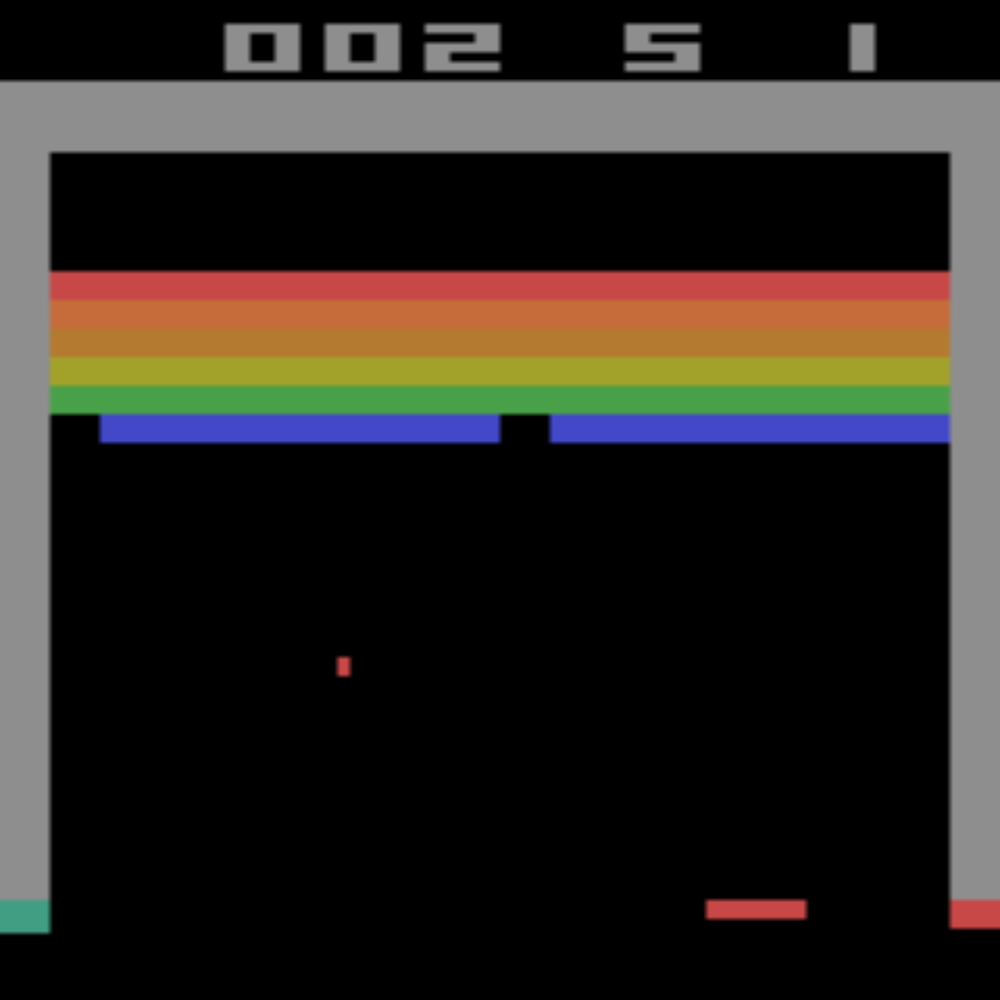}
         \caption{Breakout}
         \label{fig:breakout}
     \end{subfigure}
     \hfill
     \begin{subfigure}[b]{0.5\columnwidth}
         \centering
         \includegraphics[width=\columnwidth]{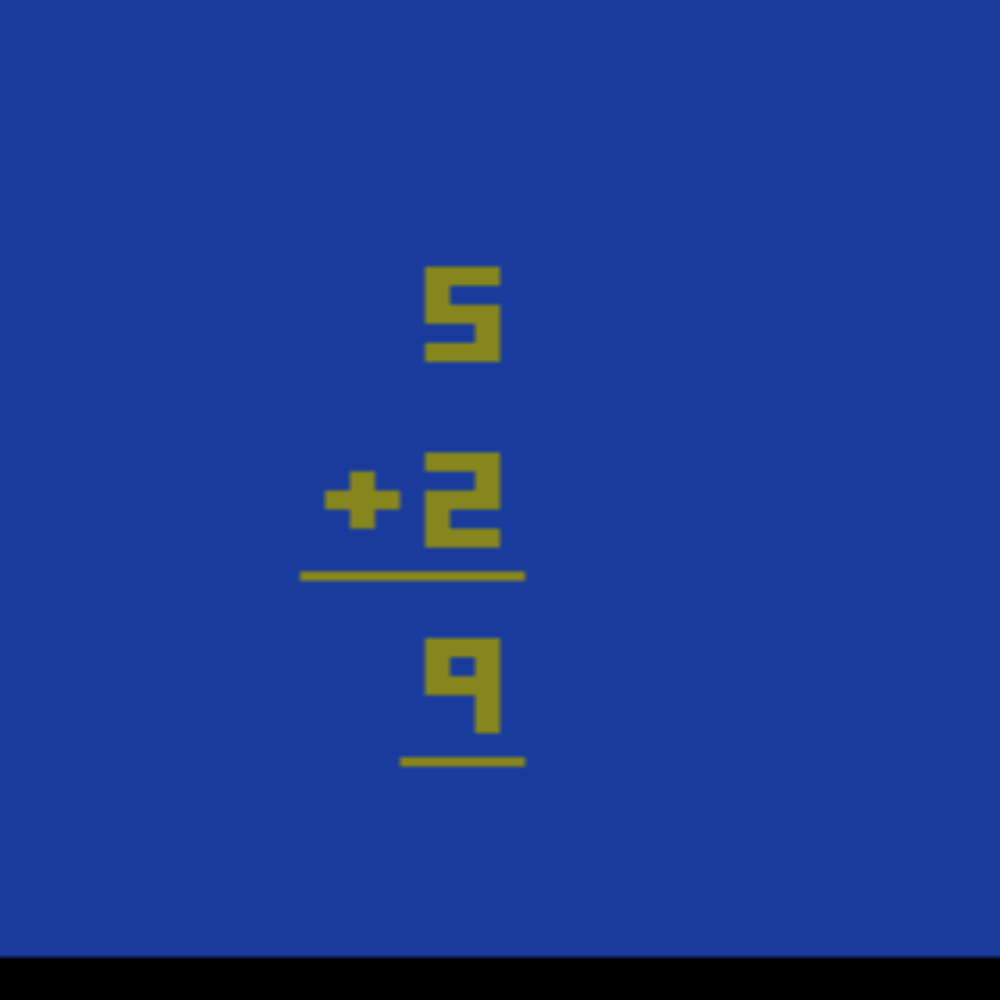}
         \caption{Basic Math}
         \label{fig:basicmath}
     \end{subfigure}
    \hfill
    \begin{subfigure}[b]{0.5\columnwidth}
         \centering
         \includegraphics[width=\columnwidth]{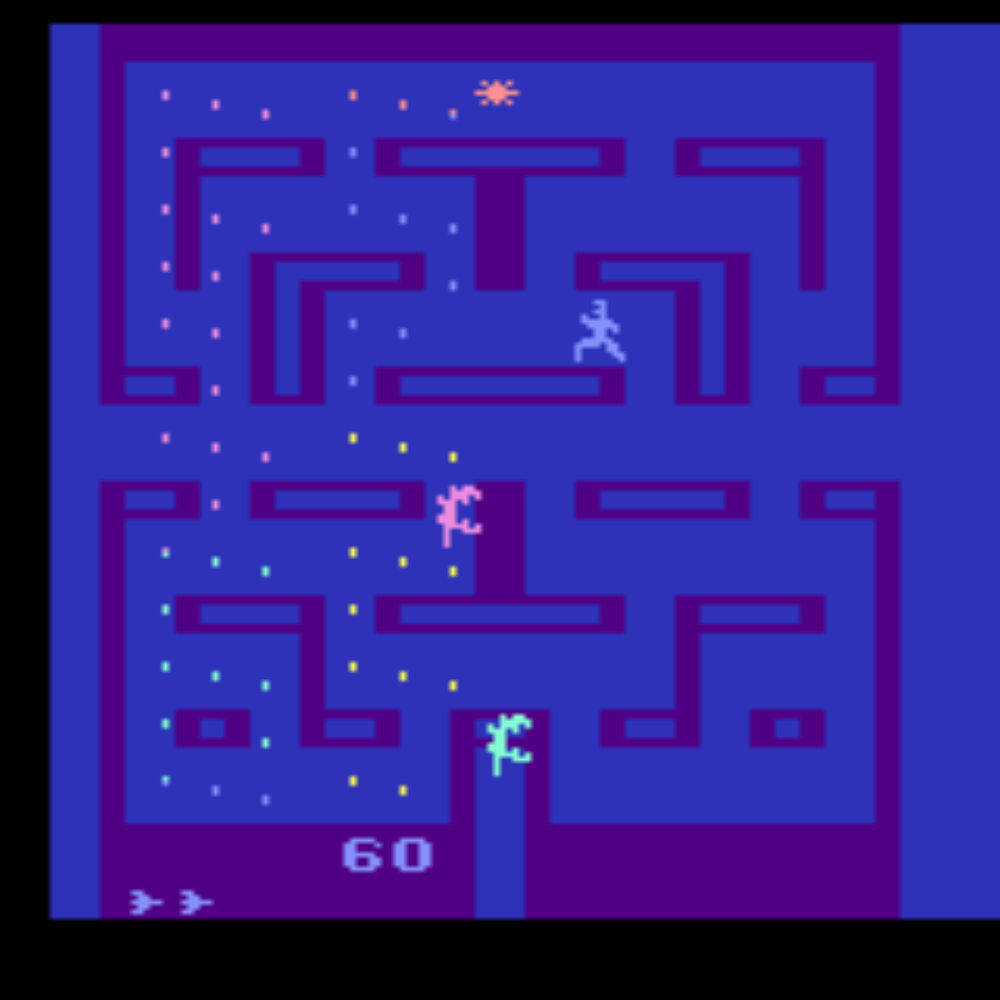}
         \caption{Alien}
         \label{fig:alien}
     \end{subfigure}
     \hfill
     \begin{subfigure}[b]{0.5\columnwidth}
         \centering
         \includegraphics[width=\columnwidth]{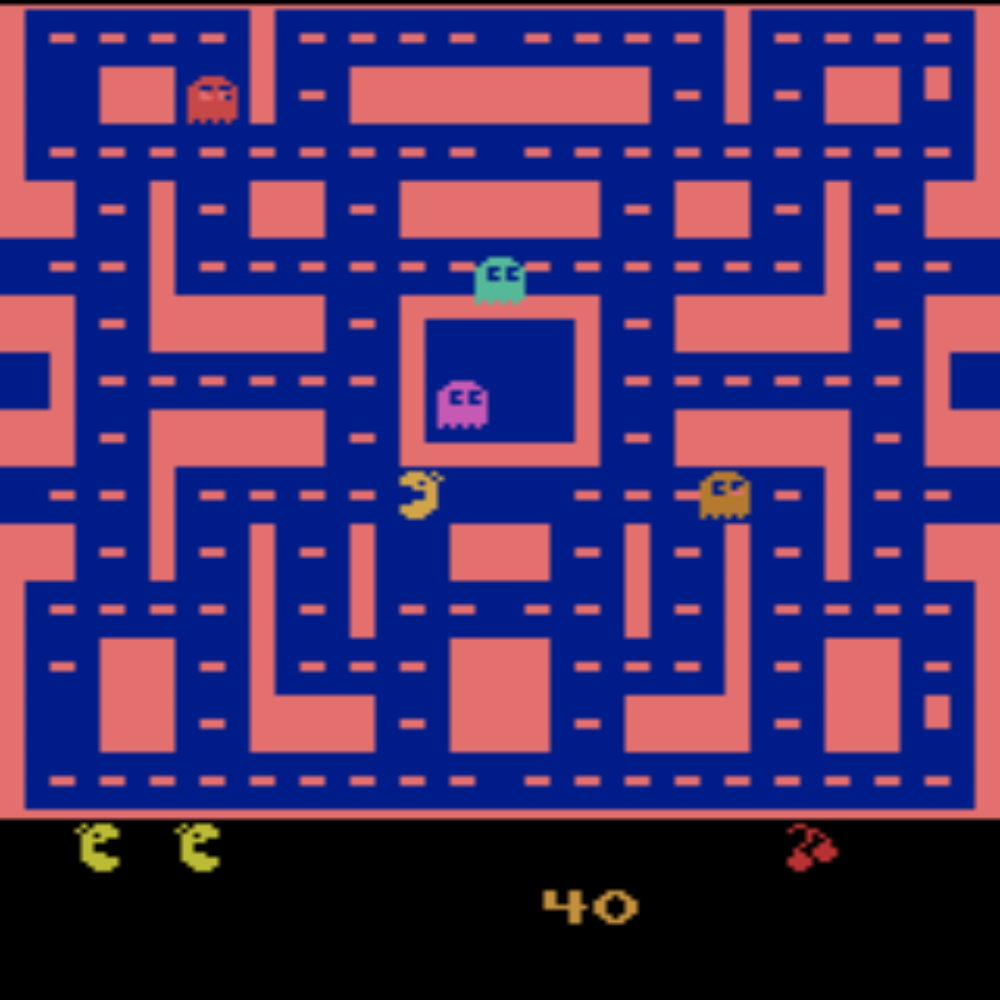}
         \caption{Ms. Pacman}
         \label{fig:ms_pacman}
     \end{subfigure}
     \hfill
     \begin{subfigure}[b]{0.5\columnwidth}
         \centering
         \includegraphics[width=\columnwidth]{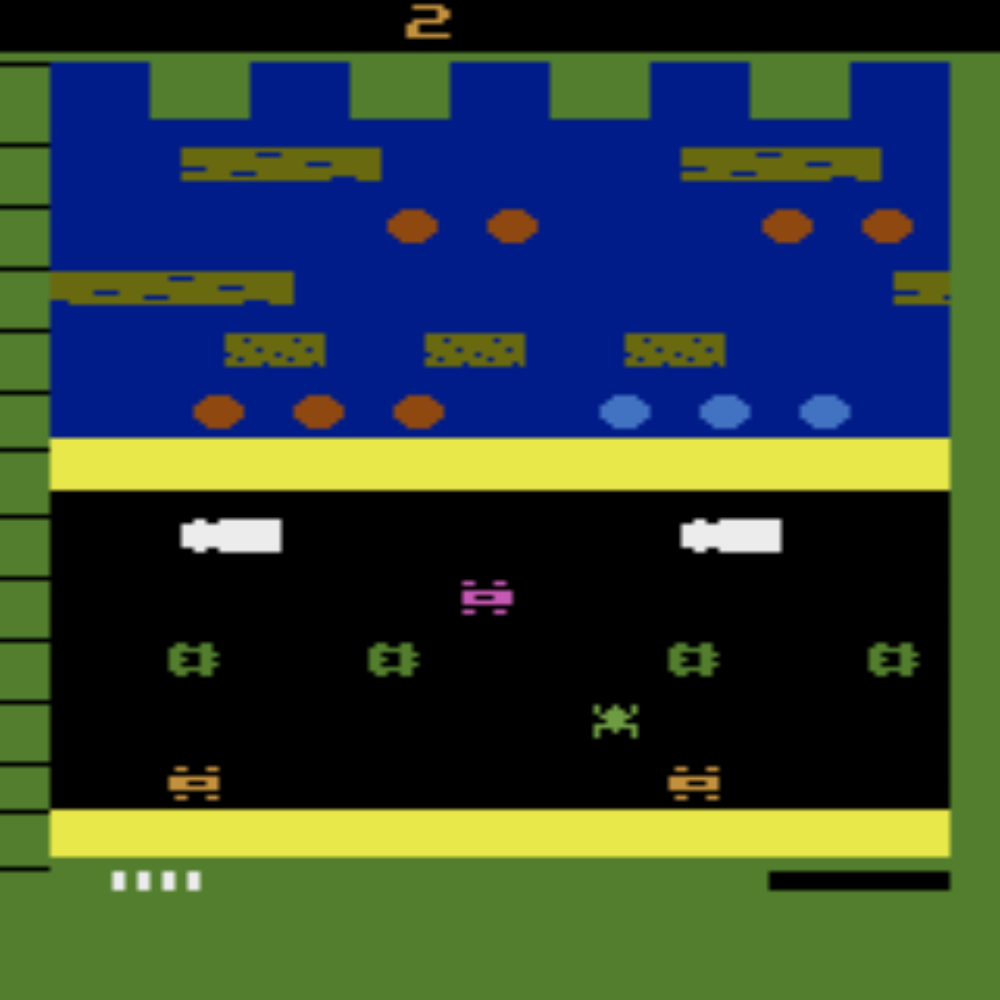}
         \caption{Frogger}
         \label{fig:frogger}
     \end{subfigure}
    \hfill
     \begin{subfigure}[b]{0.5\columnwidth}
         \centering
         \includegraphics[width=\columnwidth]{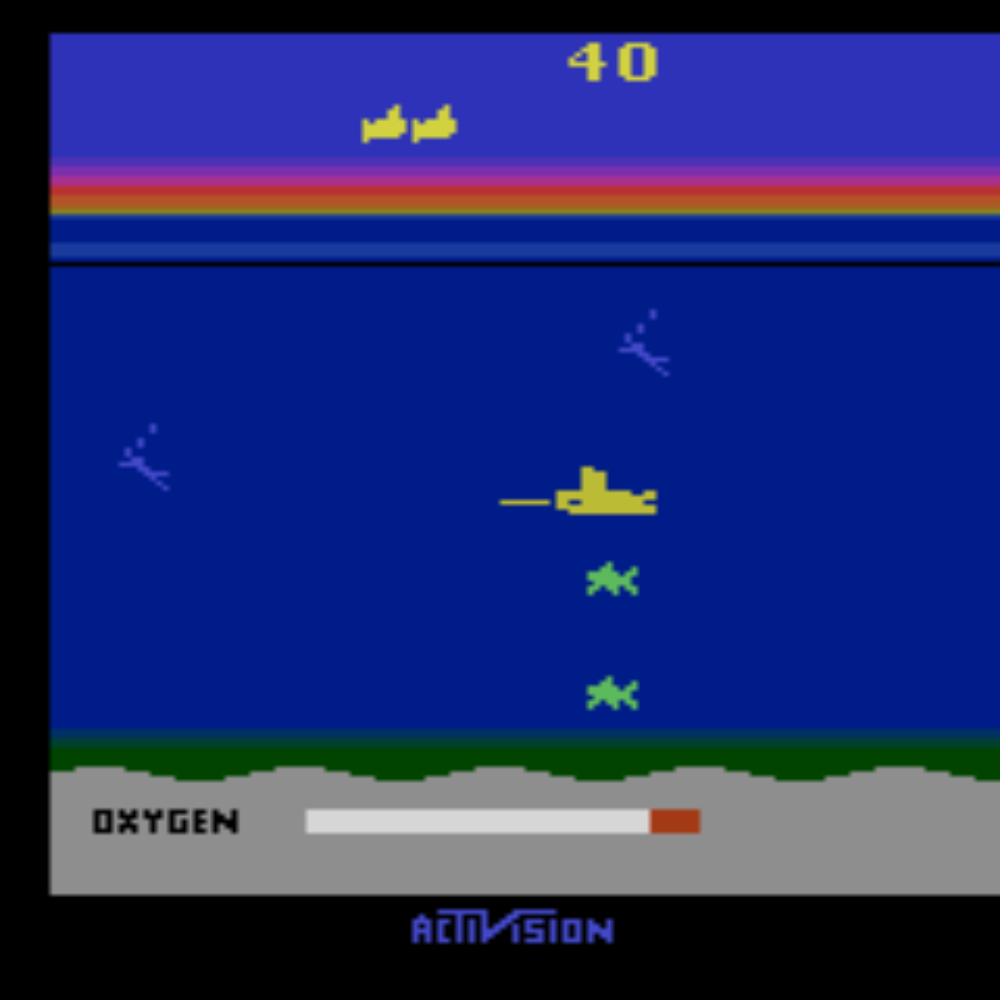}
         \caption{Seaquest}
         \label{fig:seaquest}
     \end{subfigure}
     \hfill
     \begin{subfigure}[b]{0.5\columnwidth}
         \centering
         \includegraphics[width=\columnwidth]{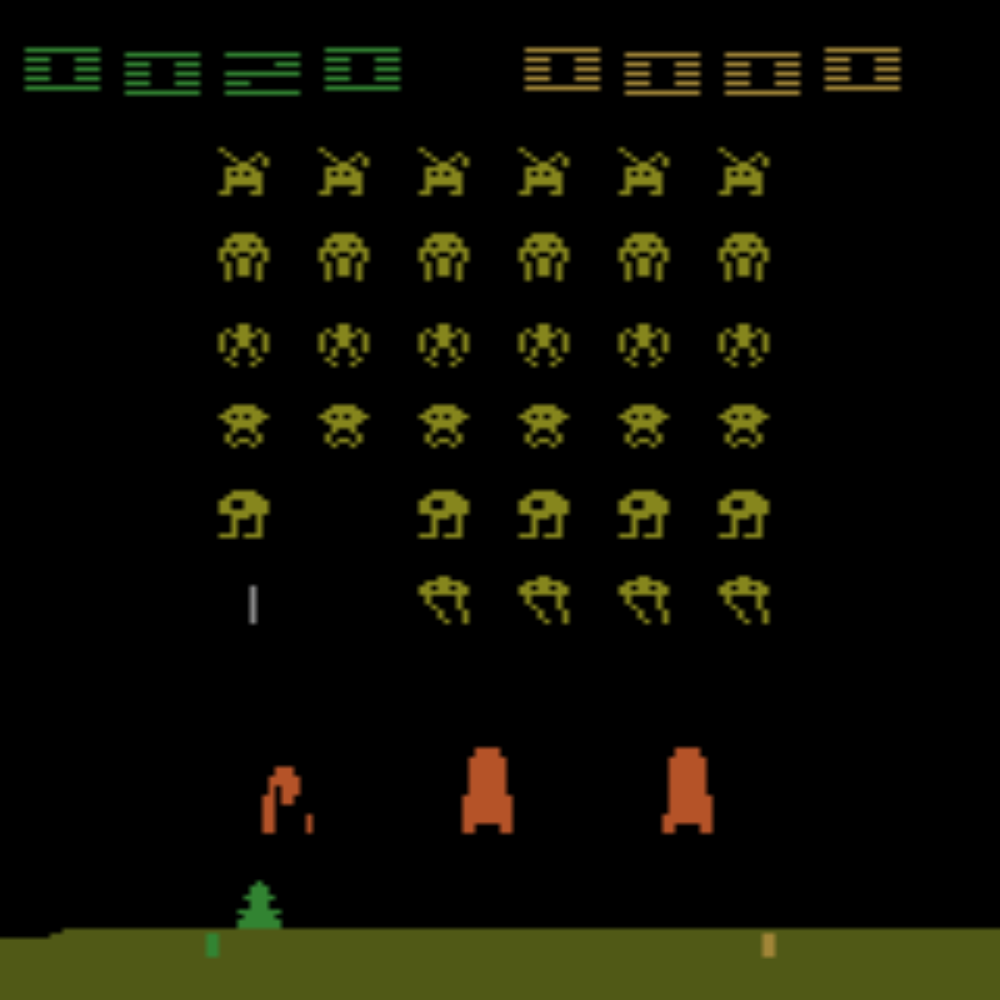}
         \caption{Space Invaders}         \label{fig:space_invaders}
     \end{subfigure}
        \caption{Images used in Understanding tasks}
        \label{fig:frames_understanding}
\end{figure*}

\subsection{Game-Play Experiment}

We conducted experiments using GPT-4V Turbo, GPT-4o, Gemini 1.5 Flash and Claude 3 Haiku. We chose these models because GPT-4V is considered state-of-the-art performance among the largest frontier LLMs at the time of writing this paper. GPT-4o, Gemini 1.5 Flash, and Claude 3 Haiku were selected for their quicker inference speed, an important feature for real-time decision-making as a low-level policy. In our tests, the average inference time along with the API call for GPT-4o, Gemini 1.5 Flash, and Claude 3 Haiku was within 2-3 seconds, while GPT-4 Turbo had an inference time of 5-7 seconds. 

We evaluated the performance in seven Atari games from the Arcade Learning Environment (ALE)~\cite{bellemare13arcade}: Space Invaders-v4, Breakout-v4, Seaquest-v4, Pong-v4, ALE/Alien-v5, Ms. PacMan-v4, and ALE/Frogger-v5. In these experiments, the current game state was presented to the LLM, which then generated an action to be executed within the Atari environment. These models were used as low-level policies, similar to how a reinforcement learning policy, such as Deep Q-Networks (DQN)~\cite{mnih2013playing}, would act in the environment.

We create a system prompt such that the output from the model is given in a JSON format with two keys, a reasoning key containing the reasoning for why the model took an action and an action key that contains the numerical action the model would like to take:
\begin{center}
    
\begin{lstlisting}[label={lst:player_action}]
{   "reasoning": "The player character is currently located at the bottom of the screen, near an exit. The closest enemy is directly in front, one tile up, and could be threatening if no action is taken. The best course of action is to fire upwards to eliminate the threat and ensure the path remains clear.",   "action": 10 }
\end{lstlisting}

\end{center}

This is an example from the environment ALE/Alien-v5 from GPT-4o. This format was used to encourage chain-of-thought reasoning to improve the game-playing performance of the LLM~\cite{wei2023chainofthought}. The system prompt was used to maintain consistency in the structure of the output and instruct the model to be a game-play assistant. In addition, each of the system prompts was tuned by providing the LLM with the official documentation description of each of the Atari environments, specifically giving the model the action names and numerical values, as detailed in the Appendix.

Since not every frame needs to be given an action and inferencing LLMs is computationally intensive, we extend the normal frame skipping of 4 frames in ALE~\cite{bellemare13arcade} to be 8 frames. With this new frame skipping we then conduct a rollout of 1,000 timesteps, where at each step, the model is provided a context buffer of the two previous frames and responses, together with the current frame. For the rollout there may be a terminal condition met when the environment is reset, which results in the reward being carried to the next episode.
This is done because Atari does not have terminal conditions based on a number of timesteps and we wanted to maintain consistency across the results.

Additionally, each frame generated is initially of size 210x160x3 but resized to 512x512x3 for all models.
We also introduced error-handling code for cases when the model responds with an invalid action, automatically replying to the LLM to correct its error.

\subsection{Visual And Spatial Reasoning}
\label{sec:atarigpt_Understanding}
We also investigated the LLMs' capability to understand and reason based on a game frame image. We evaluated the following models: GPT-4V Turbo, GPT-4o, Gemini 1.5 Flash, Claude 3 Haiku, Gemini Pro 1.5, Gemini Ultra 1.0, Gemini Pro 1.0, Claude 3 Opus, and Claude 3 Sonnet.
This allowed us to assess the state of each frontier LLM and compare their performance across different model types and sizes, and across eight environments, as shown in Figure \ref{fig:frames_understanding}.
All models were tested using their respective web interfaces.

We created a set of prompts to investigate the models' visual reasoning, spatial reasoning, strategic intuition, and ability to identify the environment:
\begin{itemize}
    \item \textbf{Visual Understanding}: Identify all the key elements in this image. Be specific. Use at most 100 words.
    \item \textbf{Spatial Reasoning}: Where are the key elements located relative to each other? Be specific with respect to their position in the image. Use at most 100 words.
    \item \textbf{Strategy}: The given image is a screenshot of a game. Describe the ideal next move if you were playing this game. Be specific. Use at most 100 words.
    \item \textbf{Identification}: Knowing that the image came from an Atari game, identify its name. Be specific.
\end{itemize}
To quantitatively evaluate the performance of the model outputs, we created a rubric outlining the basic answers to the proposed questions, as seen in the Appendix (Table \ref{tab:ground_truth_questions}).
Given that there are several acceptable actions and strategies, we do not directly define a single correct action or plan for each state. In cases where we investigate the acceptable strategy, we rather evaluate it as either a direct action or strategy/plan that does not put the agent in harm. Harm includes losing a life or losing points within a game. 

For each environment, we resize the original frame from 210x160x3 to 1000x1000x3 and query the LLM together with the visual reasoning prompt. Once a response was received, we sent the spatial reasoning prompt, followed by the strategic and identification prompts, respectively. After receiving all outputs, we compared the multimodal LLMs' output with the rubric, resulting in a percent score for that environment. We repeated this for all environments and computed the average score over four different trials.

\section{Results}
\label{sec:results}

\subsection{Game-Playing Performance}
\label{sec:Game_Playing_Performance}
We evaluate GPT-4V Turbo, GPT-4o, Gemini 1.5 Flash, and Claude 3 Haiku across seven Atari environments and compare their scores to a random agent, trained reinforcement learning agent, and human. For each model, we perform four rollouts of 1,000 timesteps and average their cumulative reward. We then normalize this average cumulative reward against the human scores, resulting in a normalized cumulative reward that relates the LLM scores to the human scores.

\begin{figure}[ht]
\begin{center}
\centerline{\includegraphics[width=\columnwidth]{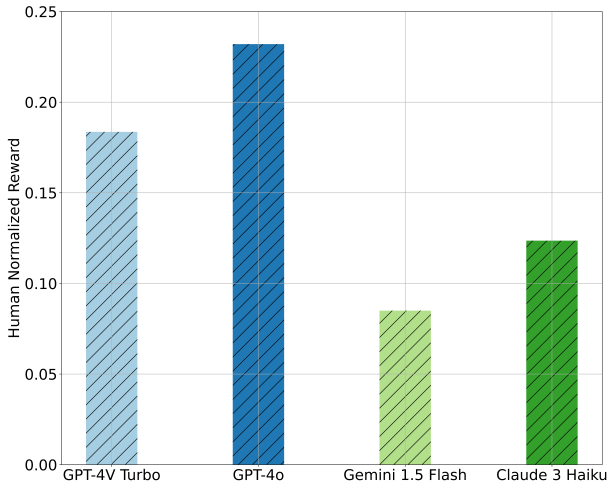}}
\caption{Normalized Average Reward for GPT-4V Turbo, GPT-4o, and Gemini 1.5 Flash.}
\label{fig:No_ICL}
\end{center}
\end{figure}

\begin{figure}[ht]
 \centering
 \includegraphics[width=\columnwidth]{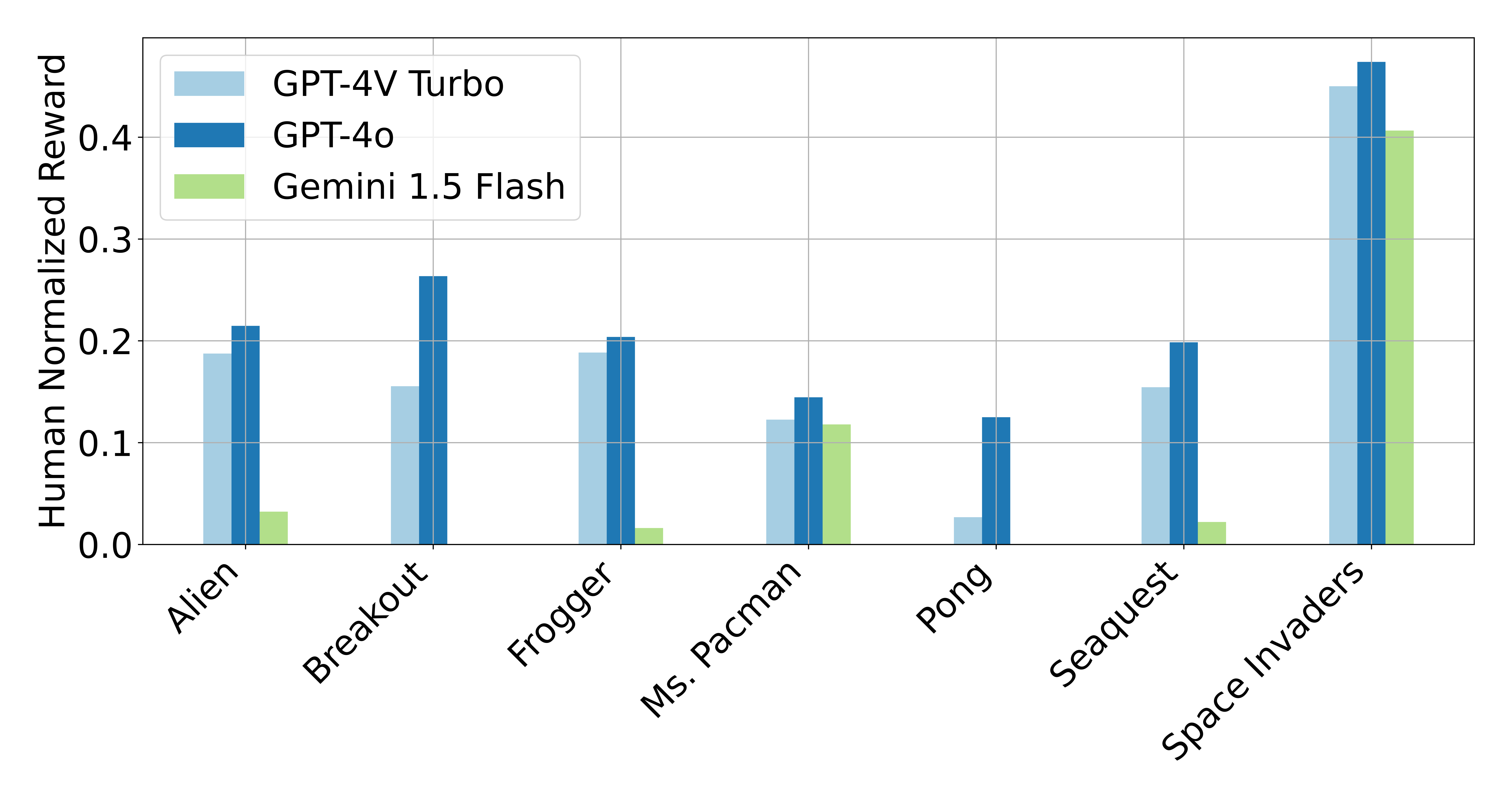}
 \caption{Average Human Normalized reward for each environment.}
 \label{fig:No_ICL_ind}
\end{figure}

As seen in Figure~\ref{fig:No_ICL}, GPT-4o performed the best on average with a normalized performance of 23.2\% and Gemini 1.5 Flash performed the worst on average with a normalized performance of 8.5\%. GPT-4V Turbo presented the second-best performance with a normalized score of 18.36\%, and Claude 3 Haiku had a normalized performance of 12.36\%. Figure \ref{fig:No_ICL_ind} breaks down the normalized reward for each environment, illustrating that the most challenging game for the LLM-based policy was Pong.


\begin{table*}[h]
\caption{Cumulative Reward for 1000 steps without In-Context Learning, * - Custom DQN model trained for 1,000,000 timesteps}
\begin{center}
\begin{tabular}{ |p{2cm}||p{1.3cm}|p{1.3cm}|p{1.3cm}|p{1.3cm}|p{1.3cm}|p{1.3cm}|p{1.3cm}|}
 \hline
 \textbf{Environments} & \textbf{Random Agent} & \textbf{RL Agent} & \textbf{Human}& \textbf{GPT-4V Turbo }& \textbf{GPT-4o}& \textbf{Gemini 1.5 Flash} & \textbf{Claude 3 Haiku}\\
 \hline
 Frogger & 26& 30* & \textbf{325}& 61.25 & 66.25 & 5.25   & 46.5\\
 \hline
 Breakout & 3 & 23  & \textbf{37} & 5.75  & 9.75 & 0   & 3.25\\
 \hline
 Pong & -20  & -8  & \textbf{2} & -25.25& -22.5 & -26 & -26  \\
 \hline
 SpaceInvaders & 100 & \textbf{725}  & 575 & 258.75 & 272.5& 233.75 & 197.5\\
 \hline
 Seaquest & 80 & 620  & \textbf{680} & 105 & 135 & 15 &40 \\
 \hline
 Alien & 270 & 1670 & \textbf{2480} & 465 & 532.5& 80 & 305 \\
 \hline
 Ms. Pacman & 280 & 3780 & \textbf{4220} & 517.5 & 610 & 497.5  &395\\
 \hline
\end{tabular}
\end{center}
\label{tab:1000_step_new}
\end{table*}
Table \ref{tab:1000_step_new} presents the raw game-play performance of the four LLMs across the Atari environments. This table also includes the performance of human players, pre-trained Deep Q-Network (DQN) reinforcement learning models~\cite{gogianu2022agents}, and random agents.
While a pre-trained DQN model\cite{gogianu2022agents} trained for 49,750,000 steps was used for all other environments, a custom DQN model was trained from scratch for 1,000,000 timesteps for ALE/Frogger-v5 due to the lack of a pre-trained model.
The LLMs did not match the performance of the human players or the RL agents. However, they outperformed the random agents, demonstrating a meaningful level of understanding and ability to play the games. This is an important finding, as it indicates that the LLMs are not merely generating random actions but are making decisions that reflect a basic comprehension of the game mechanics. Sample videos for all rollouts are available in the project webpage\footnote{Atari-GPT project webpage: \url{https://sites.google.com/view/atari-gpt/}.}.



\subsection{Visual And Spatial Reasoning}

We further explored the factors influencing game-play performance by testing the visual, spatial, strategic, and game environment identification abilities of these LLMs. For each environment, we evaluated GPT-4V, GPT-4o, Gemini 1.5 Flash, and Claude 3 Haiku using four designed prompts, which provided insight into why the models may not have performed as well as low-level policies.

\begin{figure}[h!]
\begin{center}
\centerline{\includegraphics[width=\columnwidth]{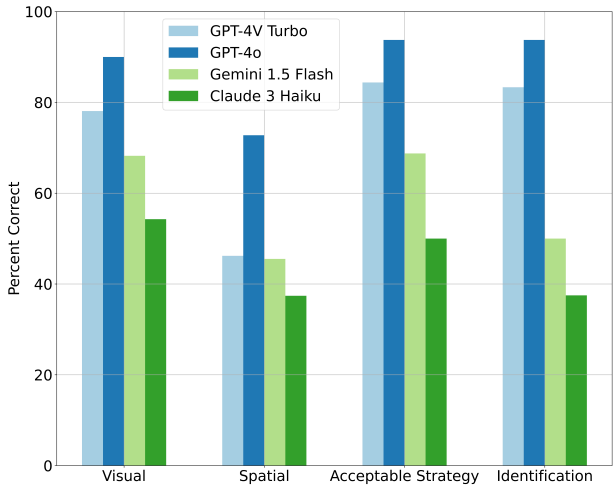}}
\caption{Visual, spatial, strategic and identification results. Percent average for 2 runs.}
\label{fig:comp}

\end{center}
\end{figure}

Figure \ref{fig:comp} displays the percentage of correct outputs for each of the four tasks—visual, spatial, acceptable strategy, and identification—across two runs for each model. GPT-4o consistently excelled across all tasks, demonstrating high accuracy in visual understanding, strategy formulation, and environment identification. However, it exhibited a noticeable decline in spatial reasoning accuracy. This pattern was consistent across all models, suggesting that spatial reasoning remains a significant challenge for multimodal large language models and possibly accounting for their relatively poor performance on the game-playing tasks. 
Comprehensive results for each environment and all models can be found in the Appendix.



\section{Discussion}
\label{sec:discussion}

This study represents one of the first attempts at benchmarking the emergent capability of multimodal LLMS to act as low-level controllers in Atari game environments, a significant departure from their traditional applications in language and visual tasks. The results, while not meeting the performance levels of human players or dedicated reinforcement learning (RL) models, showcase the potential and limitations of LLMs in this context.

Our experiments demonstrate that while LLMs exhibit some ability to identify and interact with key elements within game frames, their performance as low-level controllers is subpar, likely due to a lack of training for this task as well as difficulty in spatial reasoning. We observed a significant performance gap between GPT-4o and Claude 3 Haiku and Gemini 1.5 Flash. In most cases, we observed that models performed better than random. Though we saw performance worse than random for Pong on all models, likely due to the speed and accuracy requirements to properly play the game, and in multiple environments for Gemini 1.5 Flash, likely due to the size of the model. We observed neither large nor small models are capable of acting as zero-shot low-level controllers. While large models can comprehend the visual content fairly well, they struggle to convert this to spatial reasoning, which makes choosing a correct action more difficult. This error compounded over 1,000 frames resulted in poor performance when compared to a human player. 

Throughout our testing, we found another key element to be inference time. For these models to realistically be used for game-play tasks they will not only need to be able to see an image, interpret, and provide a correct action, but they will need to be quick enough for real-time decision-making. Our experiments show that these multimodal models still lack enough speed for acting as real-time low-level policies, as Gemini 1.5 Flash was the best in terms of inference time with an average inference taking roughly 2 seconds. 

A challenge we encountered was the inconsistency of the model's outputs, with GPT-4V Turbo occasionally failing to generate appropriate responses coupled with the above-mentioned inference time of 5-7 seconds to inference. In addition, rate limits for OpenAI, Anthropic, and Google APIs contributed heavily to much longer experimentation time, adding more overhead to the inherent inference time of these models. The imposed rate limits currently make it impossible to run real-time experiments, highlighting the need for better and faster local multimodal LLMs for fast-paced, low-level decision-making tasks.

\section{Conclusions}

Despite these setbacks, the findings are invaluable for several reasons. First, they contribute to our understanding of the current emergent capabilities and boundaries of LLMs when applied to low-level control tasks. Second, they offer a new benchmark for the AI research community to measure the progress of LLMs in handling dynamic and visually complex environments. Adjustments such as tuning the models' temperature settings demonstrated some mitigation of output inconsistency, suggesting pathways for refining LLM performance in these tasks.

Importantly, the continuous updates to LLM architectures and training methods suggest that the capabilities of these models will evolve, potentially overcoming some of the current deficiencies noted in our study. As such, this research should be viewed as a foundational step that sets the stage for future investigations, encouraging ongoing refinement and adaptation of LLMs for applications requiring detailed environmental interactions and decision-making.

While LLMs have not yet reached the level of proficiency required to match the best human or RL performances in Atari gameplay, their ability to engage in this task at all is notable. It demonstrates the adaptability and potential of LLMs to extend beyond their original training confines, offering a glimpse into future emergent applications where these models could serve as more general low-level controllers.



\section{Related Work}
\label{sec:relatedwork}

\subsection{Multimodal Large Language Models}

Processing multimodal inputs such as images and sequential data has undergone constant evolution in the domain of deep learning.
Before the transformer architecture~\cite{vaswani2023attention}, Convolutional Neural Networks (CNNs)~\cite{lecun1998gradient,krizhevsky2012imagenet} for visual processing and Recurrent Neural Networks (RNNs)~\cite{mikolov2010recurrent} for handling sequential data such as text or audio represented the state of the art~\cite{mao2015deep}.
Data was processed through separate input networks and their latest outputs were combined via different fusion strategies~\cite{mao2015deep}.
Despite achieving notable success, these approaches were limited in their scale and capacity to capture the intricate interactions between different modalities, primarily due to the inherent limitations in sequential data processing and cross-modal synthesis~\cite{chung2019sensor}.

The advent of transformers introduced a more effective and scalable mechanism for processing sequential data through self-attention mechanisms~\cite{vaswani2023attention}.
Among the key developments was the creation of CLIP (Contrastive Language-Image Pre-training)~\cite{radford2021learning}, which leveraged transformers to learn a common latent space for both visual and linguistic data, leading to a model that could correlate images in the context of natural language.
This development led to some of the most influential Multimodal Large Language Models available today such as GPT-4 Vision~\cite{openai2024gpt4}, Gemini Pro 1.5~\cite{reid2024gemini}, Gemini Ultra and Pro 1.0~\cite{geminiteam2024gemini}, Ferret~\cite{you2023ferret}, Vicuna~\cite{chiang2023vicuna}, Claude 3~\cite{Claude3}, Multimodal Large Language and Vision Assistant~\cite{liu2023visual} and LLaVa~\cite{liu2023visual}.
Since then, multimodal LLMs have been applied to different domains such as designing reward functions~\cite{ma2023eureka} and controlling general game-playing agents~\cite{SIMA}.

\subsection{Multimodal LLMs as Low-Level Policies for Games}

Low-level policies act as controllers, processing observations from the environment and returning actions.
The accessibility and complexity of games make them ideal benchmarks for evaluating the performance of such policies~\cite{mnih2013playing,badia2020agent57}.
Traditionally, video game-playing policies have employed reinforcement learning algorithms~\cite{mnih2013playing}, behavior cloning~\cite{hussein2017imitation}, or a combination of both~\cite{goecks2019integrating}.
Given the increased performance of multimodal LLMs, they have emerged as an alternative to these methods.

The rationale for employing multimodal LLMs as low-level policies in gaming is grounded in their distinctive capabilities and how they align with the demands of various game environments.
When playing social games against one another, LLMs perform well when playing games that require valuing their self-interest but sub-optimally when they need to coordinate with other players~\cite{akata2023playing}.
When fine-tuned on gameplay data, LLMs have been shown to learn an internal representation of game states that can be used to make predictions~\cite{li2022emergent}.
Given their natural language processing capabilities, LLMs can also directly learn from human-written game manuals to accelerate learning and improve their performance~\cite{wu2024read}.

Several works have demonstrated the capabilities of LLMs when playing games.
Gato~\cite{reed2022generalist} leverages a transformer architecture~\cite{vaswani2023attention} similar to LLMs to tokenize multimodal data from multiple tasks, including playing games and robotic control, to train a generalist policy. The same model with the same weights can then play games, caption images, control robotic arms, chat, and others.
CICERO~\cite{meta2022human} leveraged LLMs to combine strategic reasoning and natural language to cooperate, negotiate, and coordinate with other players to play the game Diplomacy at a human level.
LLMs have also been employed to solve text-based games~\cite{yao2020keep,tsai2023can} and directly write code to convey more complex behaviors when solving open-ended tasks in Minecraft~\cite{wang2023voyager}.

While the applications of LLMs in gaming have demonstrated considerable success across a variety of contexts~\cite{gallotta2024large}, a comprehensive exploration of these multimodal capabilities remains unexplored.
In this work, we address this gap by specifically investigating their visual, spatial reasoning, and strategic capabilities when playing Atari games.

\section{Acknowledgements}

This research was sponsored by the Army Research Laboratory and was accomplished under Cooperative Agreement Number W911NF-23-2-0072. The views and conclusions contained in this document are those of the authors and should not be interpreted as representing the official policies, either expressed or implied, of the Army Research Laboratory or the U.S. Government. The U.S. Government is authorized to reproduce and distribute reprints for Government purposes notwithstanding any copyright notation herein.

\bibliography{aaai25} 

\newpage
\appendix
\onecolumn
\section{Appendix}

\subsection{Gameplaying Performance Prompts}
\label{sec:game_Prompts}

\begin{itemize}
    \item \textbf{Alien.} ``You are a game playing assistant and will be provided an image. This will be of the game Alien, your goal is to provide me with what you believe to be the best action I could take to beat the game. Think about all possible actions and why each action is or is not the best action to take. The potential actions I can take are '0' NOOP, '1' FIRE, '2' UP, '3' RIGHT, '4' LEFT, and '5' DOWN, '6' UPRIGHT, '7' UPLEFT, '8' DOWNRIGHT, and '9' DOWNLEFT,’10’: UPFIRE, ’11’: RIGHTFIRE, ’12’: LEFTFIRE, ’13’: DOWNFIRE, ’14’: UPRIGHTFIRE, ’15’: UPLEFTFIRE, ’16’: DOWNRIGHTFIRE, ’17’: DOWNLEFTFIRE. Provide output as a json structured as {reasoning: reasoning for actions and why to choose an action, action: The environment action which would provide the best next state}. The action key should only have the action I should take for the current frame as a number."

    \item \textbf{Breakout. }``You are a game playing assistant and will be provided an image. This will be of the game Breakout, your goal is to provide me with what you believe to be the best action I could take to beat the game. Think about all possible actions and why each action is or is not the best action to take. The potential actions I can take are ‘0’: NOOP ‘1’: FIRE ‘2’: RIGHT ‘3’: LEFT. Provide output as a json structured as {reasoning: reasoning for actions and why to choose an action, action: The environment action which would provide the best next state}. The action key should only have the action I should take for the current frame as a number."
    
    \item \textbf{Frogger. }``You are a game playing assistant and will be provided an image. This will be of the game Frogger, your goal is to provide me with what you believe to be the best action I could take to beat the game. Think about all possible actions and why each action is or is not the best action to take. The potential actions I can take are ‘0’: NOOP ‘1’: UP ‘2’: RIGHT ‘3’: LEFT ‘4’: DOWN. Provide output as a json structured as {reasoning: reasoning for actions and why to choose an action, action: The environment action which would provide the best next state}. The action key should only have the action I should take for the current frame as a number."
    
    \item \textbf{Ms. Pacman. }``You are a game playing assistant and will be provided an image. This will be of the game Ms. Pacman, your goal is to provide me with what you believe to be the best action I could take to beat the game. Think about all possible actions and why each action is or is not the best action to take. The potential actions I can take are ‘0’: NOOP ‘1’: UP ‘2’: RIGHT ‘3’: LEFT ‘4’: DOWN ‘5’: UPRIGHT ‘6’: UPLEFT ‘7’: DOWNRIGHT ‘8’: DOWNLEFT. Provide output as a json structured as {reasoning: reasoning for actions and why to choose an action, action: The environment action which would provide the best next state}. The action key should only have the action I should take for the current frame as a number."
    
    \item \textbf{Pong. }``You are a game playing assistant and will be provided an image. This will be of the game Pong, your goal is to provide me with what you believe to be the best action I could take to beat the game. Think about all possible actions and why each action is or is not the best action to take. The potential actions I can take are ‘0’: NOOP ‘1’: FIRE ‘2’: RIGHT ‘3’: LEFT ‘4’: RIGHTFIRE ‘5’: LEFTFIRE. Provide output as a json structured as {reasoning: reasoning for actions and why to choose an action, action: The environment action which would provide the best next state}. The action key should only have the action I should take for the current frame as a number."
    
    \item \textbf{Seaquest. }``You are a game playing assistant and will be provided an image. This will be of the game Seaquest, your goal is to provide me with what you believe to be the best action I could take to beat the game. Think about all possible actions and why each action is or is not the best action to take. The potential actions I can take are '0' NOOP, '1' FIRE, '2' UP, '3' RIGHT, '4' LEFT, and '5' DOWN, '6' UPRIGHT, '7' UPLEFT, '8' DOWNRIGHT, and '9' DOWNLEFT,’10’: UPFIRE, ’11’: RIGHTFIRE, ’12’: LEFTFIRE, ’13’: DOWNFIRE, ’14’: UPRIGHTFIRE, ’15’: UPLEFTFIRE, ’16’: DOWNRIGHTFIRE, ’17’: DOWNLEFTFIRE. Provide output as a json structured as {reasoning: reasoning for actions and why to choose an action, action: The environment action which would provide the best next state}. The action key should only have the action I should take for the current frame as a number."
    
    \item \textbf{Space Invaders. }``You are a game playing assistant and will be provided an image. This will be of the game Space Invaders, your goal is to provide me with what you believe to be the best action I could take to beat the game. Think about all possible actions and why each action is or is not the best action to take. The potential actions I can take are ‘0’ NOOP ‘1’ FIRE ‘2’ RIGHT ‘3’ LEFT ‘4’ RIGHTFIRE ‘5’ LEFTFIRE. Provide output as a json structured as {reasoning: reasoning for actions and why to choose an action, action: The environment action which would provide the best next state}. The action key should only have the action I should take for the current frame as a number."
\end{itemize}

\newpage

\newpage

\newpage
\subsection{Ground Truth Answers for Visual and Spatial Reasoning.}

\begin{table}[ht]
\caption{Ground truth values used by human evaluators to score performance of LLMs when answering questions about game images.}
\scriptsize
\label{tab:ground_truth}
\centering
\begin{tabular}{ |p{3cm}||p{2.75cm}|p{5cm}|p{2.75cm}| }
 \hline
 \textbf{Environment} & \textbf{Visual} & \textbf{Spatial} \\
 \hline
 Alien & Player, 2 Aliens, orbs, some power up, score (60), lifes (2) & Player is in the center, one alien is below the center, other alien is center bottom, all orbs are on the left, score is at the bottom middle left and life's are bottom left \\
 \hline
 Basic Math & 3 numbers (5, 2, 9), addition sign, 2 horizontal lines & 5 is at the middle top, 2 is below 5, the addition sign is to the left of the 2, one horizontal line is below the 2, 9 is below that horizontal line and the other horizontal line is below the 9  \\
 \hline
 Breakout & Score (2), 5 lives, 1 (I am not sure what this is), a 6 lines of bricks with different colors, a red ball and a paddle & Score is top left, life's is top middle, 1 is top right, lines of bricks are in the center near the top of the gameplay area, red ball is middle left and paddle is bottom right \\
 \hline
 
 Frogger & Vehicles (9), life's (4), player, logs (7), leaves (10), score (2) & 9 vehicles all at the bottom half of the screen, life's is at the bottom left, the player is at the bottom right in-between vehicles, logs are on the top half of the screen, leaves are at the top half of the screen, score is at the top center\\
 \hline
 Ms. Pacman & Ms. Pacman, red ghost, blue ghost, pink ghost, orange ghost, orbs, power ups (2), life's (2), score (40), cherry& Ms. Pacman is in the center, red ghost is top left, blue ghost is middle top, pink ghost is center, orange ghost is middle right, orbs are throughout the environment, 2 power ups on top left and bottom left, 2 lifes are bottom left, score is bottom center, cherry is bottom right\\
 \hline
 Pong & 2 paddles, a ball and 2 scores (0,0) & Orange paddle top left, green paddle, middle right, ball top center, orange score top left, green score top right \\
 \hline
 Seaquest & Submarine, shot, fish (2), divers (2), oxygen, Activision, life's (2), score & Submarine is center, shot is to the left of the submarine, one fish is directly below the submarine and the other is directly below that fish, one diver is above the submarine and the other is to the center left of the screen, oxygen bar is bottom left and is almost full, Activision logo is bottom center of the screen, the life's are at the top center of the screen and the score is above and to the right of the life's \\
 \hline
 Space Invaders & Aliens (33), home base (3), player, shot, score, time(?)& Aliens are aligned in the throughout the middle of the game-play area, home bases are bottom center with one damaged, player is bottom left, shot is center left, score is top left, time is top right \\
 \hline 
\end{tabular}

\vspace{0.2cm}

\label{tab:ground_truth_questions}
\end{table}




\newpage
\subsection{Comprehensive Understanding Results}
\label{sec:Appendix_comp_results}


\begin{figure}[ht]
\begin{center}
\centerline{\includegraphics[width=\columnwidth]{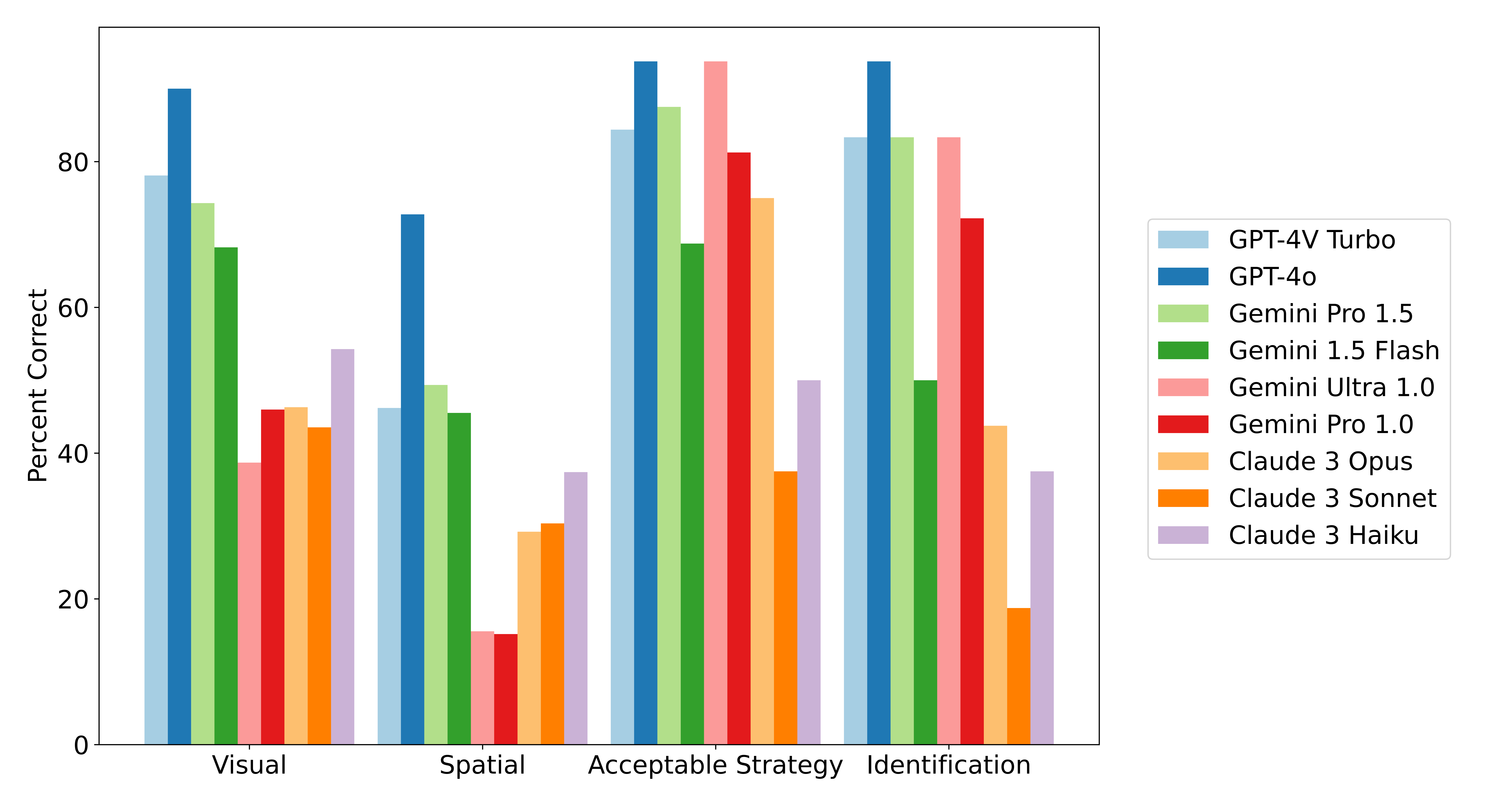}}
\caption{Comprehensive Understanding Test results.}
\label{fig:comp_appendix}
\end{center}
\end{figure}

\newpage
\subsection{Individual Performance for Visual and Spatial Reasoning}
\label{sec:Appendix_vis}


\begin{figure}[ht]
     \centering
     \begin{subfigure}[b]{0.49\columnwidth}
         \centering
         \includegraphics[width=\columnwidth]{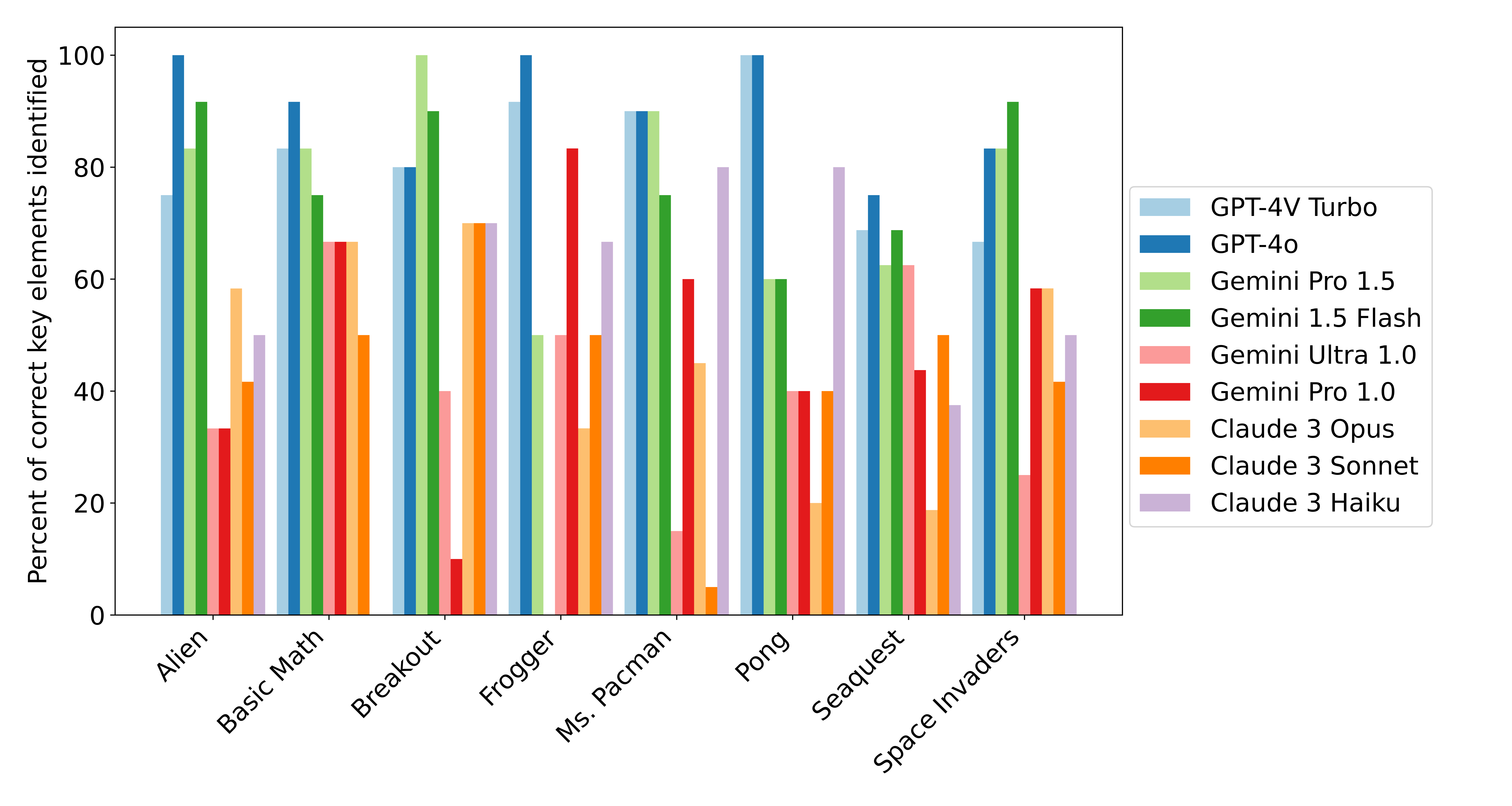}
         \caption{Visual performance}
         \label{fig:Visual_performance}
     \end{subfigure}
     \hfill
     \begin{subfigure}[b]{0.49\columnwidth}
         \centering
         \includegraphics[width=\columnwidth]{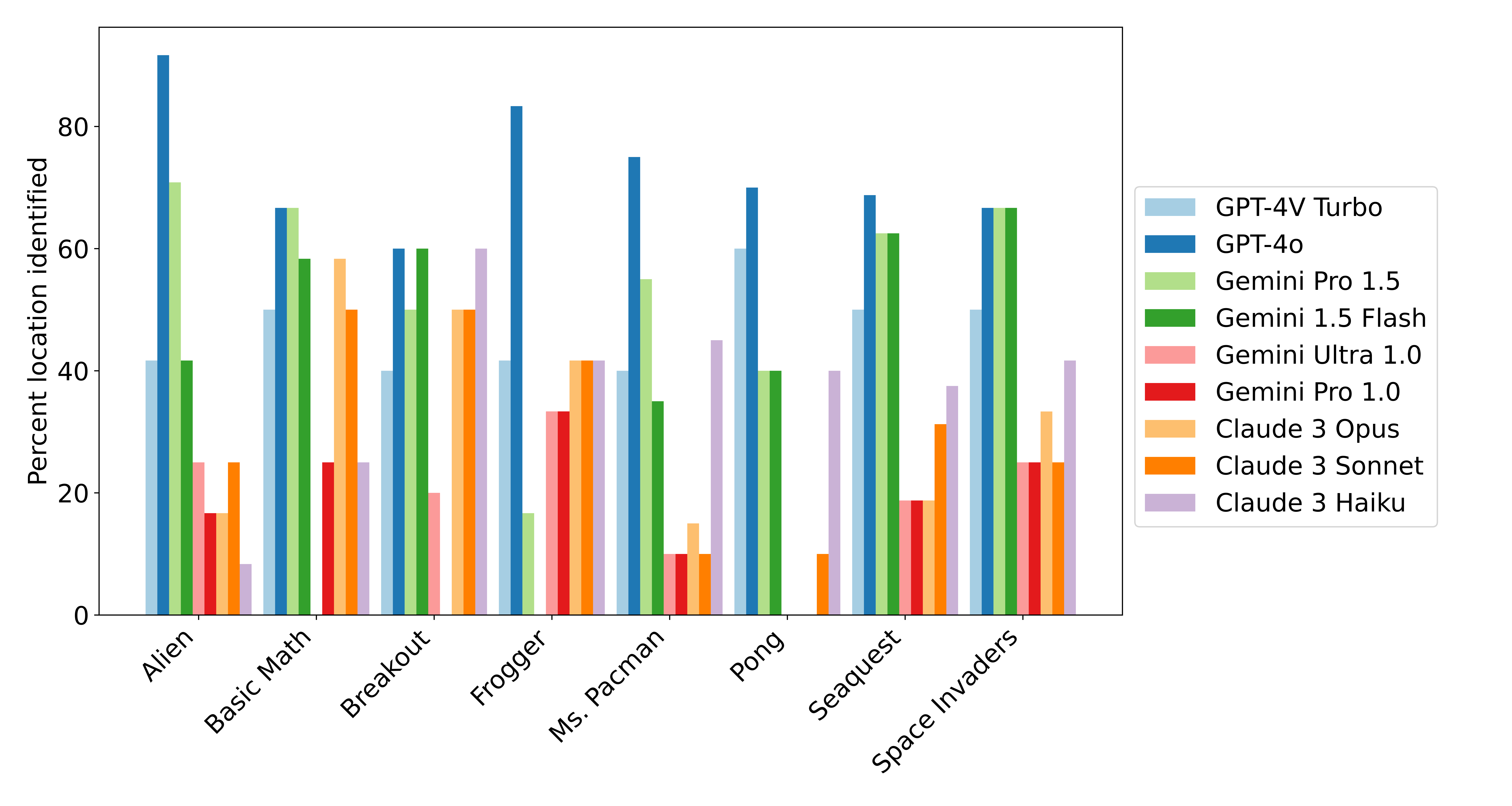}
         \caption{Spatial performance}
         \label{fig:Spatial_performance}
     \end{subfigure}
     \hfill
     \begin{subfigure}[b]{0.49\columnwidth}
         \centering
         \includegraphics[width=\columnwidth]{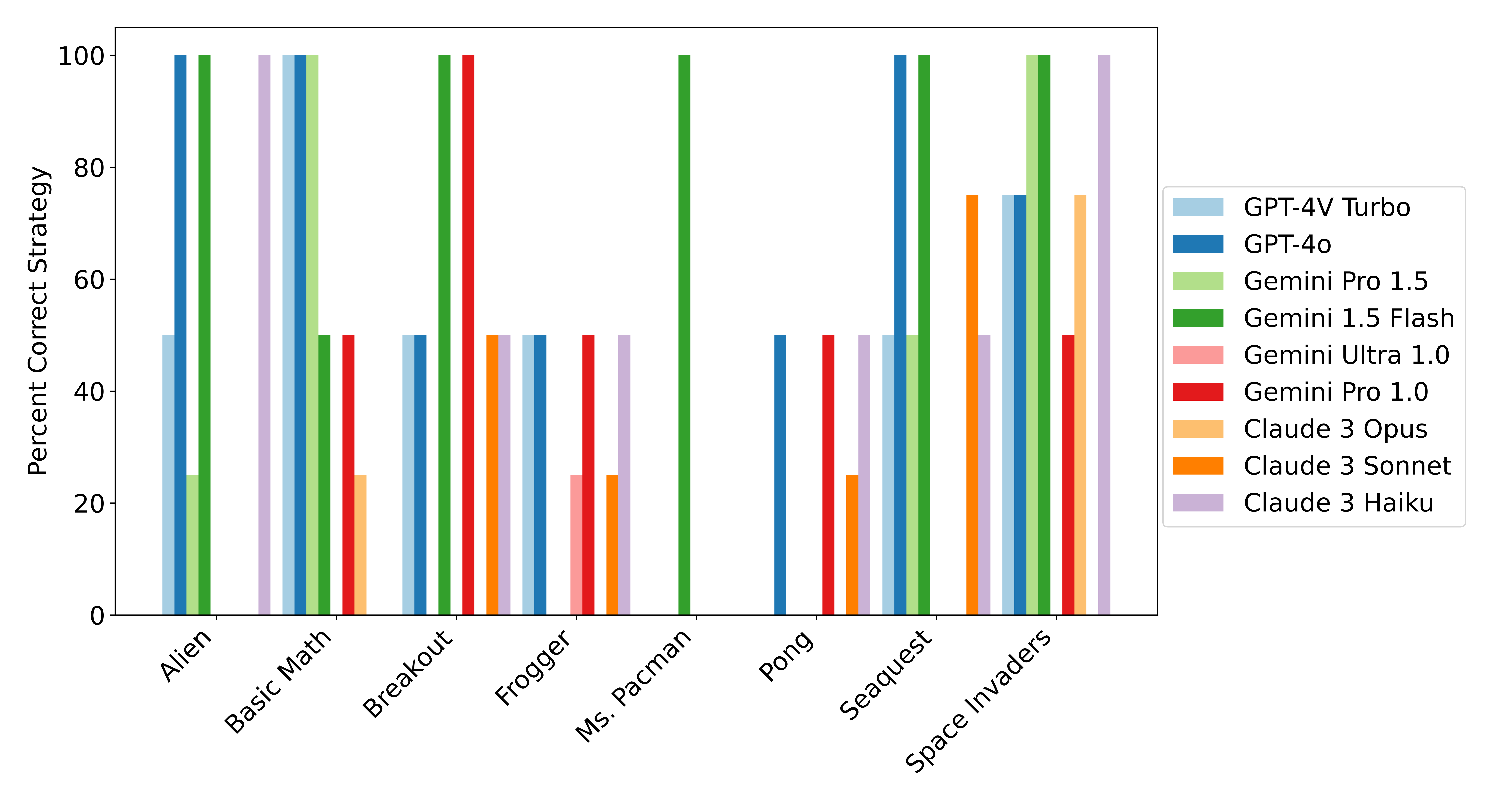}
         \caption{Strategic performance}
         \label{fig:Strat_performance}
     \end{subfigure}
     \hfill
     \begin{subfigure}[b]{0.49\columnwidth}
         \centering
         \includegraphics[width=\columnwidth]{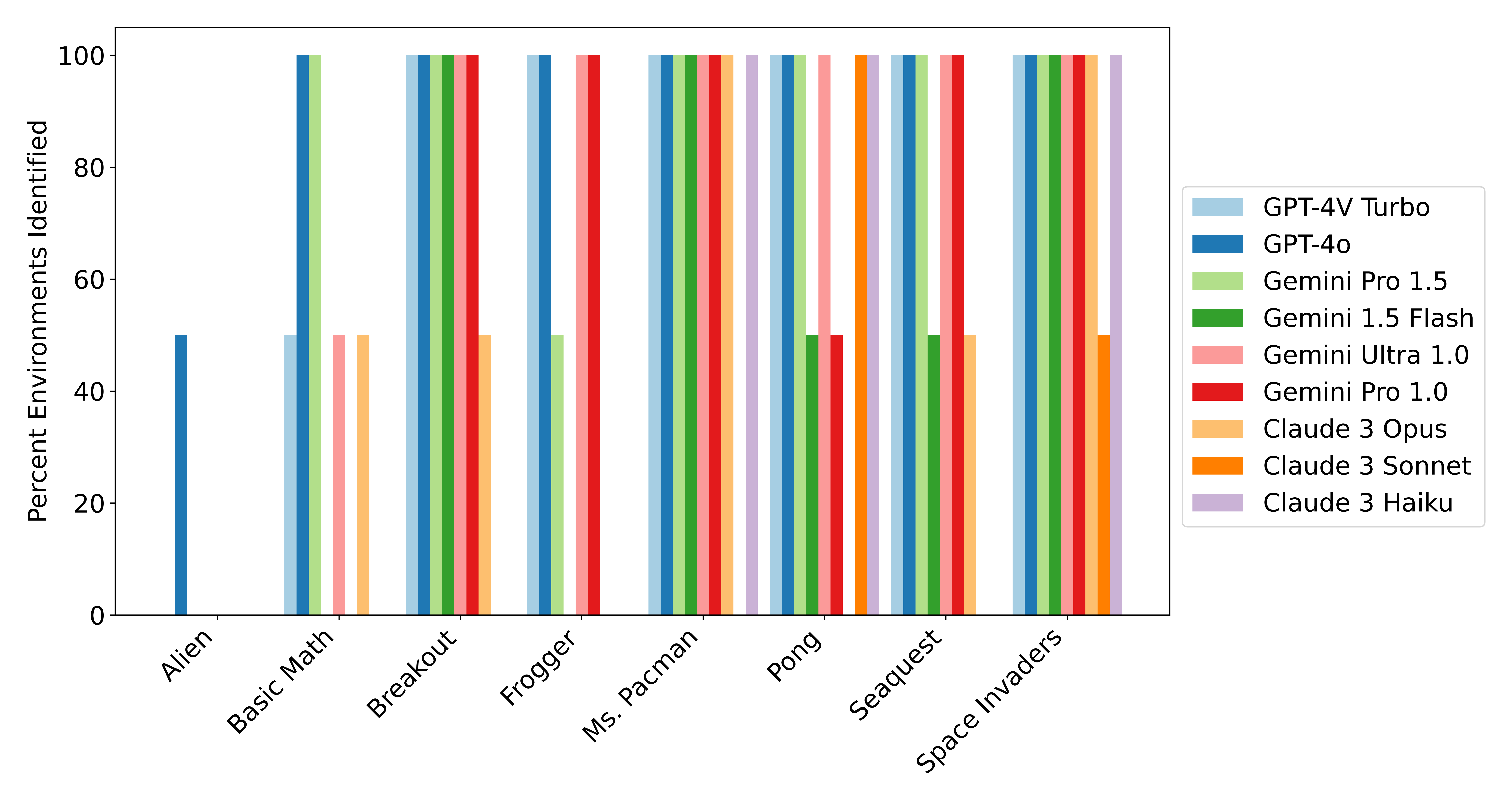}
         \caption{Identification performance}
         \label{fig:Identify_performance}
     \end{subfigure}

        \caption{Percent Performance for Individual Environments}
        \label{fig:norm_mean_reward_appendix}
\end{figure}

\end{document}